\documentstyle[psfig,epsfig]{eurospeech}

\begin{document}

\title{ 
\vspace{-2.3cm}
{\footnotesize In {\em Proceedings of
5th European Conference on Speech Communication \& Technology
(EUROSPEECH '97)}, Rhodes, Greece, 22--25 Sept 1997} 
\hspace*{\fill} \\
\vspace{1.0cm}
        Semantic Processing of Out-Of-Vocabulary Words in a 
        Spoken Dialogue System}

\author{Manuela~Boros$^1$, Maria~Aretoulaki$^2$, Florian~Gallwitz$^2$,
Elmar~N{\"o}th$^2$, Heinrich~Niemann$^2$ \and \\
        $^1$Bavarian Research Center for\\ 
Knowledge Based Systems (FORWISS)\\
        Am Weichselgarten 7\\
        D-91058 Erlangen, Germany \\
        e-mail: boros@forwiss.uni-erlangen.de
\and \\ 
        $^2$Dept.~for Pattern Recognition\\
        University of Erlangen-Nuremberg\\
        Martensstrasse 3\\
        D-91058 Erlangen, Germany\\
        e-mail: \{{\it name}\}@informatik.uni-erlangen.de}
\date{}

\baselineskip=11pt

\def\catfont{\sc}

\maketitle
\thispagestyle{empty}

\begin{abstract}
One of the most important causes of failure in spoken dialogue systems
is usually neglected: the
problem of words that are not covered by the system's vocabulary
(out-of-vocabulary or OOV words). In this paper a methodology is
described for the detection, classification
and processing of OOV words in an automatic train timetable information
system~\cite{Eckert:93}. The various extensions that had to be effected
on the different modules of the system are reported, resulting in the 
design of appropriate dialogue strategies, as are 
encouraging evaluation results on the new versions of the 
word recogniser and the linguistic processor.

\end{abstract}

\section{Introduction}

\renewcommand{\thefootnote}{\fnsymbol{footnote}}
\footnotetext[1]{Part of this work was supported by the DFG (Germany) and the
European Community.}
\renewcommand{\thefootnote}{\arabic{footnote}}
The majority of speech understanding systems have to face the
problem of words that are not covered by their current lexicon,
i.e. OOV words. In such a case the word
recogniser usually recognises one or more different words 
with a similar acoustic profile to the unknown. 
These misrecognitions often result in possibly irreparable 
misunderstandings between the user and the system. This is due to the fact
that users rarely realise that they have crossed the boundaries of
the system's knowledge but just notice its suddenly weird behaviour.
Therefore it is desireable to have the
system detect unknown words and inform the user about them so that
s/he might correct the error. This will not only increase the dialogue
success rates but also the acceptability of the system to the user
(cf.~\cite{LuperFoy:96}).

In \cite{Gallwitz:96} a method was proposed on how to integrate information
about the presence of OOV words into statistical language models.
This approach allows for both the detection of OOV words by the 
recogniser and the assignment of a semantic category to each occurrence.
In this paper, the further processing of OOV words in a spoken
dialogue system is investigated for the domain of train timetable 
inquiries~\cite{Eckert:93}. Information on OOV words 
pertaining to certain categories, as provided by the
recogniser, can be further employed by the linguistic processor and the
dialogue manager, leading to a more cooperative system.
The linguistic processor has been extended, so that it can 
integrate the information
about the occurrence of OOV words and pass on the respective semantic
data to the dialogue manager. In order for the system to react
appropriately to OOV words, special dialogue strategies have been 
devised and implemented.

First the extension of the word recogniser is described for the
detection and categorisation of OOV words.  
Then the changes and extensions effected on the linguistic processor
and the dialogue manager are sketched out. Finally,
evaluation results are reported for the modified word recogniser
and linguistic processor and the new dialogue strategies are
illustrated.

\section{Detection and Classification of OOV Words}
\label{recogniser}

In~\cite{Gallwitz:96} we presented an approach for the detection of OOV words 
which implicitly provides information on the word category. This
involves the integration of both detection {\em and} classification
of OOV words directly into the recognition process of an HMM-based word 
recogniser. 
With our approach, acoustic information as well as language model information
can be used for the purpose of classifying OOV words into different word 
categories.  Currently the same acoustic models are used for all OOV words;
only language model information contributes to the assignment of a category
to each.

\begin{figure}[tb]\centering
\mbox{\psfig{figure=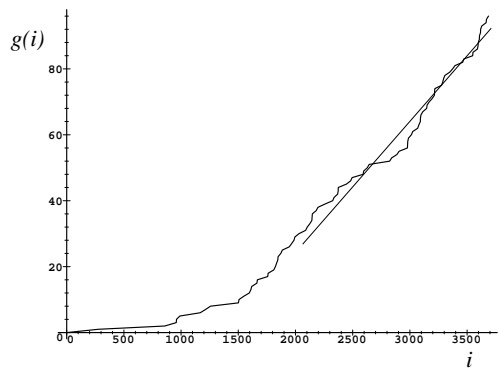,width=\columnwidth}}
\caption{Estimation of the current OOV word probability for
word category {\catfont city}. The function $g$ gives the number 
of words in category {\catfont city} up to the $i$th word of the training sample
that would have been OOV if we had redefined the vocabulary after
each observed word. The slope of the linear approximation
is an estimation of the OOV probability of category {\catfont city}
.}
\label{city_rate}
\end{figure}

The basic idea behind our approach is to build language
models for the recognition of OOV words that are based on a system of
word categories. 
Emission probabilities of OOV words are then estimated for each word category.
Even if we include in our vocabulary all words of a category that were observed 
in the training sample, there is still a certain probability of observing
other new words of the same category in an independent
test sample or in future
utterances. This probability can be estimated from the training sample itself.
Details on the calculation of the OOV emission probabilities were given
in~\cite{Gallwitz:96}. 
Figure~\ref{city_rate} shows the principle of this estimation technique
for the category {\catfont city} of our spontaneous train timetable inquiry 
sample.

For most of our linguistically-motivated word categories, the OOV probability
is 0, because they describe a finite set of words. In the
time table inquiry domain there are 5 word categories
that are practically infinite 
(e.g. {\catfont city, region, surname}).
In addition, a category for rare words has been defined that do not
fall under any other category
(OOV probability 73\%) and another for garbage (e.g. word fragments,
OOV probability 100\%).

After integrating OOV probabilities into the language model, the latter
has to be combined with one or several acoustic
models for OOV words. Simple `flat' acoustic models can be used 
for this purpose as well as more enhanced models based on 
phone- or syllable-grammars.

\section{Extensions to the Linguistic Processor}

Typically the Linguistic Processor's (LP) task in a spoken dialogue
system is to build a semantic representation of the user utterances
with the aid of the system's linguistic knowledge base (i.e. grammar
and lexicon). The semantic representation will be further processed by the 
dialogue manager in order to be assigned an interpretation 
on the basis of the actual dialogue context, which will ultimately 
guide the system's reaction. In our system, the input to the LP 
is a string of best-scored word hypotheses generated by the word 
recogniser. 
The grammar formalism used is Unification Categorial Grammar (UCG),
while for the representation of the semantic content of utterances
the Semantic Interface Language (SIL) is employed 
(\cite{McGlashan:90}). A detailed description of the
linguistic processor can be found in \cite{Meckl:95}.

Word strings delivered by the OOV--extended recogniser will contain the
respective information if an OOV word has been uttered.
In order to make this information accessible to the dialogue manager,
the LP has to be modified to include it
into the semantic representation that is passed on to the dialogue
manager. The system will then be capable of reacting appropriately to
an OOV word.

In Section \ref{recogniser} the categories that can 
actually be assigned by the recogniser were introduced. 
For the domain of train timetable inquiries, categories
such as {\catfont surname} or {\catfont garbage} are not relevant for the
proper understanding
of user utterances and can, hence, be neglected in later processing. 
However, it is most desireable to retain any information that unknown
city names were uttered, as city names constitute a central part of 
the domain. In order for the LP to handle OOV words of category 
{\catfont city}, an appropriate lexical entry had to be included in its
lexicon. Syntactically this lexical entry is equivalent to the
specified city names, semantically it differs in that it simply
carries the information that the name of the corresponding city 
is unknown to the system. 
The respective feature
structure-like entry is shown below in an abbreviated and mnemonic form.

\begin{verbatim}
morphology: form: oov_city,
syntax: head: (part_of_speech: proper_noun,
               number:singular),
semantics: (type:location,
            thecity:( type:city,
                      value:oov_city)).
\end{verbatim}

The slot {\tt semantics} contains the semantic specification of a sign.
{\tt Type:location} means that the sign denotes a certain location that
is further specified by the role {\tt thecity}, which carries 
the information that it is a city whose name is defined by {\tt
value}. Given that an unknown city name is involved here, the 
respective value is {\tt oov\_city}.

The addition of the {\it oov\_city} entry guarantees that an input
string containing a word of type {\it oov\_city} can properly be
parsed and its semantic representation correctly built up, leading
to the following representation:

\begin{verbatim}
semantics: (type: go,
            thegoal: (type: city,
                      value: oov_city)).
\end{verbatim}

This semantic representation correctly contains the information that
the goal of the journey specified by the user is not covered by the
lexicon. This representation is passed on to the dialogue manager
for interpretation.

\section{Extensions to the Dialogue Manager}
\label{DMan}

The role of the Dialogue Manager (DMan) in the system is to locate the
data that is relevant to the task in the semantic representations provided
by the LP, so that the train information database can be accessed.
Secondly, it is therein that the next system
utterance is planned in accordance with what the user has said and the
current state of the dialogue \cite{Eckert:96,Eckert:93}
(cf.~\cite{LuperFoy:96}).  In the case of
train timetable inquiries, there are two types of relevant semantic 
objects: the task parameters that should be specified by the user 
before the DB can be accessed, namely {\em goalcity, sourcecity, date}
and {\em goaltime} or {\em sourcetime}; and various dialogue markers 
(e.g. {\tt right}, {\tt no}, {\tt thanks}) 
which influence the user-oriented progression
of the dialogue. The most central component of the DMan is the
Dialogue Module, which keeps track of the state of the dialogue 
in terms of system and user dialogue acts, as well as system goals 
and their satisfaction. An ATN description of the possible dialogue
step transitions is used to generate expectations about the 
continuation of the dialogue, in terms of both user and system acts. 
This is also the main submodule that had to be extended in order to 
incorporate OOV word information and have appropriate system
utterances formulated accordingly
(Section~\ref{Dstrategies_examples}).  

Before the incorporation of OOV word information in the system, 
when an OOV word was uttered in relation to one of the task
parameters, the DMan would process an acoustically similar 
city name, for instance. This did not lead to an immediate 
dialogue failure, as the user was always able to correct the system 
later on, in which case the system would fall back to its default 
recovery strategies: it would start by requesting the corresponding 
parameter value again and cross-checking the remaining parameters
after the first or second repetition (and failure). Then the user
would be asked to spell the problematic word. Failure to acquire 
an utterance interpretation at this stage would force the system to 
close the dialogue by referring the user to a human information 
officer.  

The extension of the word recogniser and the LP of the system 
with meta-knowledge about the occurrence of OOV words has led to 
the design of new dialogue strategies that take this extra 
information into account and are adopted on-line in the presence of an
OOV word (Fig.~\ref{dstrategies}). Thus, two new dialogue 
states were incorporated in the corresponding ATN description,
which accommodate alternative state transitions in the DMan 
accordingly: (a) {\em repeat\_param} is used to ascertain that 
an OOV word was indeed uttered, in order to avoid false alarms. It 
provides a first warning to the user that there may be a problem 
and asks him/her to repeat just the parameter value 
involved. (b) {\em warn} follows the default repair mode {\tt spell} 
and involves the notification of the user about the cause of failure 
so that he/she can either hang up or pose a different query. 
These extensions of the DMan are illustrated in 
Section~\ref{Dstrategies_examples}.    

\section{Experiments and Results}

The evaluation experiments on the word recogniser and
the linguistic processor were performed on the EVAR
corpus collected while the system
was accessible via the German public telephone network. A total 
number of 1092 dialogues with (naive) users were recorded, comprising 10556
utterances consisting of 37775 words. As test sample we used a subset
of these 1092 dialogues containing 2383 utterances.

\subsection{Evaluation of the Recogniser} \label{WA_eval}

Experiments were carried out using a simple acoustic OOV word model that 
consists
of a fixed number of HMM states with equal probability density functions.
For a vocabulary size of 1110 words the OOV rate was 5.3\% in the test
sample. Word accuracy (WA) was evaluated by substituting all OOV words 
by the symbol OOV in both the
reference data and the recogniser output~\footnote{This
common approach does not take into account that we actually {\em classify}
OOV words. Thus, recognising ``{\em Hamburg}'' as {\tt oov\_city} is
one full recognition error. The corresponding problem is discussed in
Section~\ref{lingeval} in the context of semantic concept accuracy (CA).}.
 
In the experiments described in this paper a word error rate 
reduction of 5\% was achieved (Table~\ref{TabelleA}). The {\em Precision} 
(ratio of
correctly detected OOV words to the number of hypothesized OOV words)
was 30.7\% while {\em Recall} (ratio of the number of correctly detected
OOV words to the total number of OOV words in the reference data) was 23.7\%.
The increase in word accuracy despite the dissatisfactory Precision ratio
is due to the fact that OOV false alarms mostly occur when the baseline
recogniser produces a recognition error anyway.

Our goal is not only to {\em detect} but also to {\em classify} OOV words.
Of all {\em correctly} recognised OOV words (matches of reference OOV words
and hypothesized OOV words), the word category is assigned
correctly in 94\% of the cases. For the two-class-problem
{\catfont CITY} {\em vs.} not-{\catfont CITY} the recognition rate is 97\%.
These encouraging results show that even pure language model information
enables the word recogniser to reliably distinguish between OOV words of 
different word categories.

\subsection{Evaluation of the Linguistic Processor}\label{lingeval}

For the evaluation of the linguistic processor, the metric of semantic
concept accuracy (CA) is used. CA measures the system's
ability to detect the semantic concepts that are necessary in order 
to understand an utterance and was described in detail in \cite{Boros:96}.

In order to assess the functionality of
the extended LP alone, initial testing employed the transliterations 
of the 2383 utterances as input to the parser. The resulting figure 
of 93.8 \% shows that the semantic coverage of the system is very
good, especially if one keeps in mind that the system deals with 
spontaneous speech (even if transliterated).
For the evaluation of the word recogniser and the LP in combination,
the recogniser output is taken as input to the parser. Two
separate experiments were carried out: one without the 
possibility to detect and process OOV words and another with the 
possibility to do so.
The first experiment without OOV word information yielded a CA of 
73.2 \%, the respective WA of the recogniser being 77.1 \%. 
Extending the recogniser to accommodate the detection and 
classification of OOV words increases its WA to 78.5 \% and 
accordingly results in a higher CA rate of 75.1 \% for
the extended LP. Table~\ref{TabelleA} shows the corresponding figures 
for CA in each case:

\begin{table}[hbt]
\begin{center}
\begin{tabular}{|l||c|c|} \hline
 INPUT & WA & CA \\ \hline\hline
 transliterations & --- & 93.8 \\ \hline
 with OOV & 78.3 & 75.5 \\ \hline
 without OOV & 77.1 & 73.2 \\ \hline
\end{tabular}
\end{center}
\caption{Preliminary results of LP evaluation.}
\label{TabelleA}
\end{table}

These figures indicate that the correlation between WA and CA reported
in \cite{Boros:96} also holds in the experiments described here.
The improvement of the recogniser's WA due to OOV word detection reported
in \ref{WA_eval} also improves the linguistic processor's CA. 

These results are based on a very strict interpretation of the CA measure:
the misrecognition of a (possibly badly pronounced) city name that is
in the vocabulary, e.g. ``{\em Hamburg}'', as {\tt oov\_city} leads to 
a semantic representation
that is ``almost'' correct; the system reaction of asking the
user to repeat the particular piece of information~(see 
Section~\ref{Dstrategies_examples}) would be quite natural. 
We believe that users would be more tolerant to this specific 
kind of error. However, this counts as one ``full'' error. 
Thus, optimising the CA for the recogniser--parser combination 
will {\em not} lead to the ideal overall system performance.
Consequently, a better
measure for CA would probably be to count this type of error only as
a ``50\%--error''. This hypothesis will be further investigated when 
a sufficiently large sample of dialogues has been collected
with the OOV--extended dialog system.

\subsection{Example Dialogue Strategies for OOV Words}
\label{Dstrategies_examples}

\begin{figure}[t]
\centerline{\psfig{figure=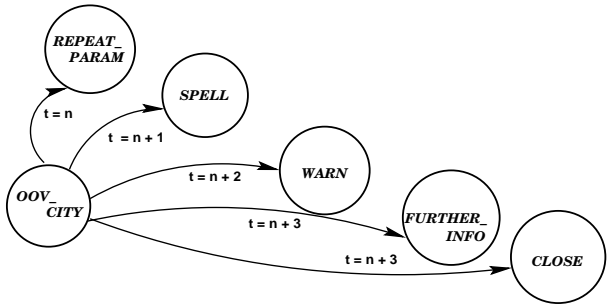,width=80mm,height=40mm}}
\caption{\label{dstrategies} Dialogue Step Transition Network 
                             for OOV words}
\end{figure}

On the basis of the user's reactions in the course of the dialogue 
and the frequency of conflict between the system's beliefs and 
the user's goals, the system can dynamically modify its communicative
and repair strategies, e.g. whether or not there is a confirmation
goal for the system~\cite{Eckert:96}. The incorporation in the DMan 
of the two new dialogue states introduced in Section~\ref{DMan}, which
also consist new system goals, has resulted in a number of
alternative dialogue step transitions, the most general of which is 
schematically shown in Fig.~\ref{dstrategies} and exemplified in 
the following dialogue. The labelling of a word as {\em OOV\_CITY} 
forces the system to postpone its current goal (e.g. request for another 
parameter) and activate a confirmation/warning sub-dialogue which should 
render the system more user-oriented and response generation more 
acceptable to the user (cf.~\cite{LuperFoy:96}).

\begin{small}
\begin{verbatim}
User:   I want to go to <Brussels> oov_city.
          [goalcity:oov_city]
System: I think the information you require is not
        covered by our database. Could you, please, 
        repeat the name of the city you want to go 
        to?  
          [system goal: REPEAT_PARAM]
User:   <Brussels> oov_city.
System: Could you please spell the name of this 
        city?
          [system goal: SPELL]
User:   <B-r-u-s-s-e-l-s> oov_city.
System: Unfortunately, there is no information on 
        train connections for the city you want. Our 
        database only covers German cities. Would 
        you like to proceed with a different query?
          [system goal: WARN]
User:   Yes.
System: What exactly would you like to know?
          [system goal: FURTHER_INFO]
\end{verbatim}
\end{small}
 
The extended version of the DMan and of the complete train
information system will be shortly made available for testing and use 
over the German public telephone network.

\section{Conclusion and Future Work}

In this paper a methodology was proposed and illustrated 
for the linguistic and semantic processing of OOV words in 
a spoken dialogue system. The necessary changes and extensions 
to the word recogniser and the linguistic processor were described
as well as appropriate new dialogue strategies that modify the system
behaviour accordingly. Evaluation
results were also reported regarding the word recogniser and 
the linguistic processor, which showed an 
encouraging increase in both word accuracy and semantic concept accuracy. The
corresponding error rates dropped by 5\% and 7\%, respectively.
Those OOV words detected by the word recogniser were
correctly classified in 94\% of the cases.

Current work includes the further improvement and evaluation of 
the word recogniser and the linguistic processor of the system. 
In addition, the newly-implemented dialogue strategies will 
be tested and evaluated under realistic circumstances by making 
the extended system version accessible via the public telephone 
network, thus also collecting more test data.

\end{document}